%% file: main.tex
\documentclass[10pt,twocolumn,letterpaper]{article}

\usepackage{amsfonts}       
\usepackage{nicefrac}       
\usepackage{multirow}
\usepackage{pifont}
\usepackage{amsmath}
\usepackage{graphicx}
\usepackage{floatrow}
\usepackage{caption}
\usepackage{setspace}
\usepackage[pagenumbers]{cvpr} 

\DeclareMathOperator*{\argmin}{arg\,min}

\definecolor{mykeywordcolor}{rgb}{0.0, 0.5, 0.0}  
\definecolor{myvarcolor}{rgb}{0.7, 0.0, 0.0}     
\definecolor{myoperatorcolor}{rgb}{0.0, 0.0, 1.0} 
\definecolor{midcolor}{rgb}{0.5, 0.5, 0.5} 
\input{preamble}

\definecolor{cvprblue}{rgb}{0.21,0.49,0.74}
\usepackage[pagebackref,breaklinks,colorlinks,allcolors=cvprblue]{hyperref}

\def\confName{CVPR}
\def\confYear{2025}

\newcommand{\methodname}{Align3R\xspace}

\title{\methodname: Aligned Monocular Depth Estimation for Dynamic Videos}

\author{
Jiahao Lu$^1$\thanks{Equal contribution} \quad Tianyu Huang$^{2*}$ \quad Peng Li$^1$ \quad Zhiyang Dou$^3$ \quad Cheng Lin$^3$ \\ Zhiming Cui$^4$ \quad Zhen Dong$^5$ \quad Sai-Kit Yeung$^1$ \quad Wenping Wang$^{6}$ \quad Yuan Liu$^{1,7}\footnotemark[2]~$
\\[0.3em]
$^1$HKUST \quad $^2$CUHK \quad $^3$HKU \quad  $^4$ShanghaiTech \quad $^5$WHU \quad $^6$TAMU \quad $^7$NTU
\\ \small{* Equal contribution. Project page: \href{https://igl-hkust.github.io/Align3R.github.io/}{https://igl-hkust.github.io/Align3R.github.io/}}
\vspace{-20pt}
}

\begin{document}
\twocolumn[{%
\renewcommand\twocolumn[1][]{#1}%
\maketitle
\includegraphics[width=\linewidth]{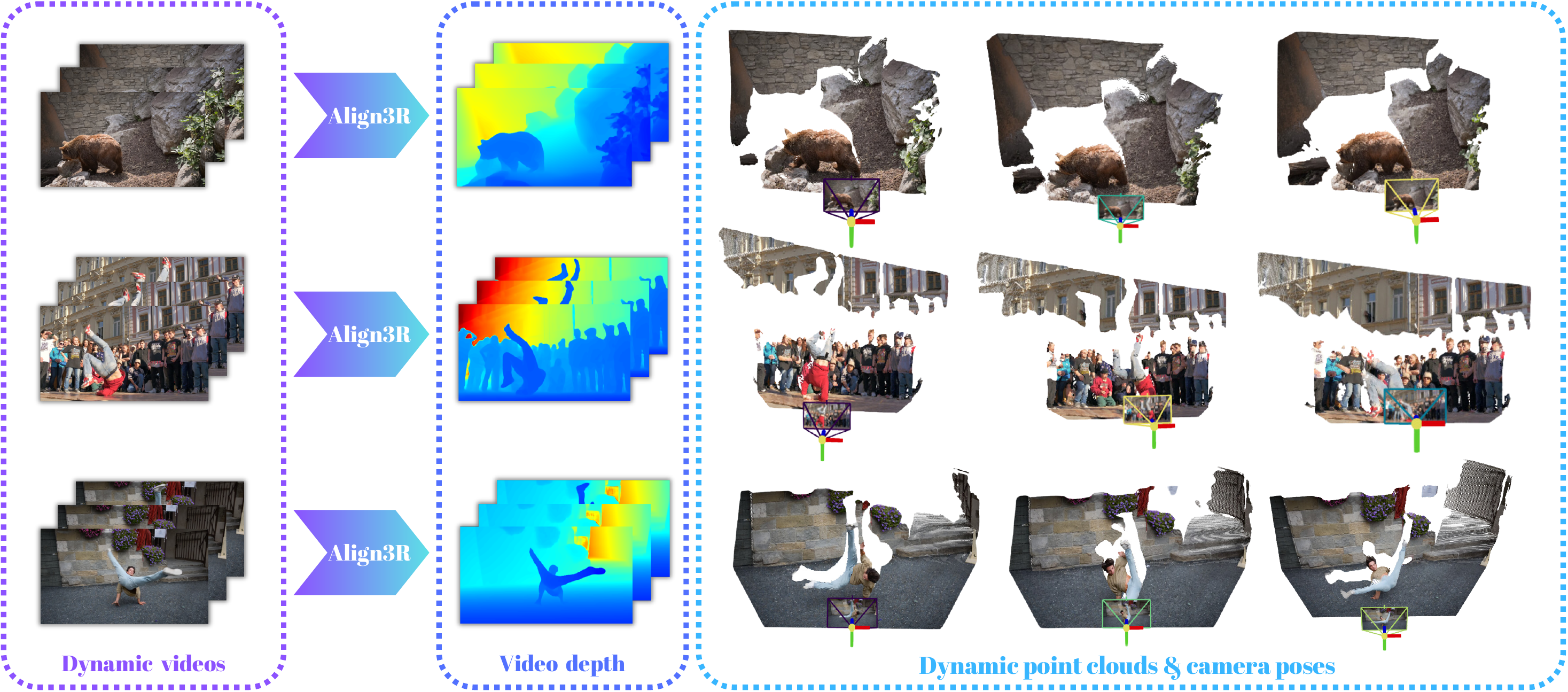}
\vspace{-2em}
\captionof{figure}{\textbf{Align3R} estimates temporally consistent video depth, dynamic point clouds, and camera poses from monocular videos.}
\label{fig:teaser}
\vspace{10pt}
}]
\renewcommand{\thefootnote}{\fnsymbol{footnote}}
\footnotetext[2]{Corresponding Author}
\input{sec/0_abstract}    
\input{sec/1_intro}
\input{sec/2_related_works}
\input{sec/3_method}

\input{sec/4_experiments}
\input{sec/5_conclusion}
{
    \small
    \bibliographystyle{ieeenat_fullname}
    \bibliography{main}
}


\input{sec/X_suppl}

\end{document}

%% file: preamble.tex
\newcommand{\myparagraph}[1]{\smallskip\noindent\textbf{#1.}}

%% file: sec/0_abstract.tex
\begin{abstract}
Recent developments in monocular depth estimation methods enable high-quality depth estimation of single-view images but fail to estimate consistent video depth across different frames. Very recent works address this problem by applying a video diffusion model to generate video depth conditioned on the input video, which is training-expensive and can only produce scale-invariant depth values without camera poses. In this paper, we propose a novel video-depth estimation method called \methodname to estimate temporally consistent depth maps for a dynamic video. Our key idea is to utilize the recent DUSt3R model to align estimated monocular depth maps of different timesteps. First, we fine-tune the DUSt3R model with additional estimated monocular depth as inputs for the dynamic scenes. Then, we apply optimization to reconstruct both depth maps and camera poses. Extensive experiments demonstrate that \methodname estimates consistent video depth and camera poses for a monocular video with superior performance than baseline methods.

\end{abstract}

%% file: sec/1_intro.tex
\section{Introduction}
\label{sec:intro}

Monocular depth estimation for monocular videos or image sequences is a crucial problem in computer vision and robotics with applications in various downstream tasks, \eg, camera localization, scene reconstruction, and so on~\cite{Hong-CVPR2022, Bian-CVPR2023, Rosinol-IROS2023}.
Traditional depth estimation methods~\cite{schoenberger2016mvs} rely on strong camera motions with a large baseline to estimate dense depth maps with stereo matching. Despite its wide applications, estimating monocular depth is challenging due to its ill-posed nature.

Recent works~\cite{yin2023metric3d,hu2024metric3dv2,bhat2023zoedepth,bochkovskii2024depthpro,yang2024depthanything,yang2024depthanythingv2,ke2024marigold,fu2025geowizard} address this ill-posed problem in a data-driven manner, which enables relative or metric depth estimation on single-view images after training on large-scale datasets. Though impressive performance is achieved on the depth estimation for a single image, estimating temporally consistent video depth for a dynamic video is still challenging. All these single-view depth estimators have difficulty maintaining a consistent scale factor of the estimated depth maps for different frames, which results in flickering artifacts in the estimated depth sequences. 

To align the depth maps of different frames, early-stage works~\cite{luo2020consistent,kopf2021robust} mainly apply inference time optimization by aligning the depth maps with the correspondence constraints across different frames built from flow estimation or image matching. However, these methods can hardly handle large dynamic motions or camera motions due to the vulnerable flow estimation or image matching and their optimization process is extremely time-consuming taking more than several hours. 
Recent concurrent works~\cite{hu2024depthcrafter,shao2024chronodepth} take advantage of powerful video diffusion models to maintain the cross-frame consistency for direct video depth generation. However, such video diffusion models often require large computation resources and dataset volume for training. Due to their computation complexity, these methods can only process video clips of a predefined length and struggle to maintain consistency across different clips~\cite{shao2024chronodepth}. Moreover, they only produce scale-invariant depth without camera poses~\cite{shao2024chronodepth,hu2024depthcrafter}, which are not sufficient for downstream tasks like 4D reconstruction~\cite{lei2024mosca,lu2024dn,liu2024modgs,wang2024shape} or 3D tracking~\cite{xiao2024spatialtracker}. Thus, estimating a consistent depth sequence from a monocular video is still challenging.

In this work, we propose a novel method called \textbf{\methodname} for consistent video depth estimation from a monocular video as shown in Fig.~\ref{fig:teaser}. The key idea of \methodname is to combine the monocular depth estimators with the recent DUSt3R~\cite{wang2024dust3r} model. Monocular depth estimators enable high-quality depth estimation for each frame with fine details but cannot maintain cross-frame consistency while the DUSt3R model can predict coarse 3D pairwise point maps to align two frames. Thus, \methodname combines the best of two models for accurate and high-quality video depth estimation, which shows the following three characteristics. First, in comparison with the video diffusion-based methods~\cite{shao2024chronodepth,hu2024depthcrafter} that generates a depth video directly, \methodname only needs to learn to predict pairwise point maps, which is much easier for learning. Second, in comparison with the original DUSt3R model, \methodname can predict more details in depth maps due to the utilization of the high-quality monocular depth estimation methods~\cite{yang2024depthanythingv2,bochkovskii2024depthpro}. Third, \methodname further naturally supports camera pose estimation of dynamic videos, which is challenging for traditional Structure-from-Motion (SfM) pipelines~\cite{schoenberger2016sfm}. The estimated camera poses further support various downstream tasks like dynamic novel-view synthesis.

Combining the monocular depth estimation with DUSt3R is non-trivial. 
A straightforward way is a post-optimization strategy that utilizes the pairwise point map prediction of DUSt3R as constraints to directly align the monocular depth estimation from different frames as early-stage works~\cite{luo2020consistent,kopf2021robust}. However, we find that this leads to sub-optimal results even after fine-tuning the DUSt3R on the dynamic videos. Instead of this post-optimization strategy, we propose a better combination strategy that utilizes the monocular depth estimation in fine-tuning DUSt3R. This pre-combination strategy effectively makes the fine-tuned DUSt3R model aware of the monocular depth maps to be aligned. Specifically, we unproject the estimated monocular depth maps to get 3D point maps, then apply an additional Transformer to encode the point maps, and finally add the encoded features into the decoder of DUSt3R. 
In this strategy, we inject the monocular depth estimation into the point map prediction of the DUSt3R model in fine-tuning, which not only results in more detailed point map prediction but also makes the fine-tuned model informed about the scale difference of input depth maps. After that, we only need to follow the optimization process of DUSt3R to predict the depth maps of different frames.

We have conducted comprehensive experiments on 6 synthetic and real-world dynamic video datasets. The results show that \methodname is able to estimate temporal consistent video depth maps and outperforms all baseline methods by a large margin. Meanwhile, our method is also able to estimate accurate camera poses for the dynamic videos, showing better or comparable performance as our concurrent work MonST3R and pose estimation methods.

%% file: sec/2_related_works.tex
\section{Related works}
\label{sec:rw}

\myparagraph{Monocular Depth Estimation}
Traditional methods~\cite{hoiem2007recovering, liu2008sift,schoenberger2016mvs,li2023opal} for depth estimation generally rely on explicit feature matching and epipolar constraints, thus having difficulty dealing with complex and low-textured scenes.
In recent years, deep learning-based methods~\cite{yang2021transformer, patil2022p3depth, yuan2022neural, bhat2021adabins, yao2018mvsnet} have presented superior performance and become mainstream. However, their generalization ability to unseen domains could be challenging due to the limited training data. To address this problem, some methods are proposed to employ affine-invariant loss functions~\cite{ranftl2020towards} for training on large-scale mixed datasets and achieve impressive accuracy and robustness regarding scale-shift-invariant relative depth estimation~\cite{yang2024depthanything, yang2024depthanythingv2,he2024lotus,garcia2024fine}. Another stream of methods focuses on directly learning metric depth by leveraging camera parameters or careful network designs~\cite{bhat2023zoedepth, piccinelli2024unidepth, yin2023metric3d, hu2024metric3dv2, bochkovskii2024depthpro}. Besides, with the success of diffusion-based generative models~\cite{rombach2022high}, new attempts are made to leverage the diffusion priors for high-quality and zero-shot depth estimation~\cite{ke2024marigold,fu2025geowizard}.
Nevertheless, all these methods focus on static monocular images and struggle to maintain cross-frame consistency for video depth estimation.

\myparagraph{Video Depth Estimation}
To achieve consistent video depth estimation, early-stage approaches generally employ inference time optimization that takes camera poses or optimal flow as constraints to align the depth maps across different frames~\cite{chen2019self, kopf2021robust, luo2020consistent, zhang2021consistent, wang2022less, xu2024depthsplat}. The performance of these methods largely depends on the accuracy of vulnerable camera pose or flow estimation, therefore struggling to deal with large camera motions or dynamic components. 
Recently, some feed-forward methods are proposed to directly predict depth sequences from videos~\cite{li2023temporally, Teed2020DeepV2D, NVDS, NVDSPLUS, yasarla2023mamo,yang2024depthanyvideo}. While achieving impressive accuracy, their generalization ability to open-world videos with diverse content could be constrained by the limited model capacity.
In addition, some methods are developed to utilize the powerful video diffusion models to directly generate high-quality video depth~\cite{hu2024depthcrafter,shao2024chronodepth}. Despite the impressive performance, their video diffusion models often require large computation resources for training. 
Due to the computation complexity, these methods can only process video clips with a predefined length.
In contrast, our method \methodname only needs to learn to predict pairwise point maps, which is much easier to learn and more flexible.


\myparagraph{Cocurrent Work}
MonST3R~\cite{zhang2024monst3r} is a concurrent work that also finetunes DUSt3R on dynamic videos for both video depth and camera pose estimation, which achieves very impressive results. The major difference is that \methodname is initially motivated by consistent video depth estimation so \methodname incorporates an estimated monocular depth in the DUSt3R model while MonST3R directly finetunes the original DUSt3R model without any additional modules. 
Other concurrent works~\cite{wang20243d,ye2024no} also adopt a DUSt3R-like framework for poseless Gaussian Splatting in static scenes.

%% file: sec/3_method.tex
\section{Method}

\textbf{Overview}. Given a video consisting of $N$ frames $\{\mathbf{I}_k\in\mathbb{R}^{\rm{H\times W\times 3}}|k=1,...,N\}$, our target is to estimate the corresponding depth maps $\{\mathbf{D}_k\in \mathbb{R}^{\rm{H\times W}}\}$ and camera poses $\{\mathbf{\pi}_k\in \mathbb{SE}(3)\}$. \methodname achieves this by first applying a monocular depth estimator, e.g. Depth Pro~\cite{bochkovskii2024depthpro} and DepthAnything V2~\cite{yang2024depthanythingv2}, to estimate depth maps $\{\mathbf{\hat{D}}_k\}$ for all frames. Then, we apply a modified DUSt3R model~\cite{wang2024dust3r} to predict pairwise point maps for some frame pairs in the video using both the image $\mathbf{I}_k$ and the estimated depth map $\mathbf{\hat{D}}_{k}$. Finally, we solve for globally consistent depth maps $\{\mathbf{D}_k\}$ and camera poses $\{\mathbf{\pi}_k\}$ for all frames using these predicted pairwise point maps. In the following, we first introduce the DUSt3R~\cite{wang2024dust3r} model.

\subsection{Recap of DUSt3R}

Given a set of images $\{\mathbf{I}_k\}$ of a static scene, DUSt3R~\cite{wang2024dust3r} can estimate depth maps and camera poses for all frames by pairwise point map prediction and global alignment.

\textbf{Pairwise prediction}. In the point map prediction, DUSt3R applies a ViT-based network that takes a pair of image frames $\mathbf{I}_n$, $\mathbf{I}_m$ $\in \mathbb{R}^{\rm{H\times W}
\times 3}$ as inputs, and outputs point maps of both frames $\mathbf{X}^{e}_n$, $\mathbf{X}^{e}_m$ $\in \mathbb{R}^{\rm{H\times W}
\times 3}$ with respect to the coordinate of frame $n$, where $e=(m,n)$, and the corresponding confidence maps $\mathbf{C}^e_n,\mathbf{C}^e_m$ $\in \mathbb{R}^{\rm{H\times W}}$.
For the given image set $\{\mathbf{I}_k\}$, DUSt3R constructs a connectivity graph $\mathcal{G}(\mathcal{V},\mathcal{E})$ for selecting pairwise images, where the vertices $\mathcal{V}$ represent $N$ images and each edge $e \in \mathcal{E}$ is an image pair.

\textbf{Global alignment}. After predicting all the pairwise point maps, DUSt3R introduces a global alignment optimization to solve for the depth maps $\mathbf{D}:=\{\mathbf{D}_k\}$ and camera poses $\pi:=\{\pi_k\}$ by
\begin{equation}\label{eq:DUSt3R_global}
     \argmin_{\mathbf{D},\mathbf{\pi},\sigma} \sum_{e \in \mathcal{E}} \sum_{v \in e} \mathbf{C}_v^{e} \left\Vert \mathbf{D}_v - \sigma_e P_e(\pi_v, \mathbf{X}_v^{e}) \right\Vert_2^2,
\end{equation}
where $\sigma = \{\sigma_e\}$ are the scale factors defined on the edges, $P_e(\pi_v, \mathbf{X}_v^{e})$ means projecting the predicted point maps $\mathbf{X}^e_v$ to view $v$ using poses $\pi_v$ to get a depth map. 
The objective function in Eq.~\eqref{eq:DUSt3R_global} explicitly constrains the geometry alignment between frame pairs, therefore after the optimization process, cross-view consistency can be maintained in the estimated depth maps.

\textbf{Motivation}.
Since DUSt3R enables the estimation of globally consistent depth maps while monocular depth estimators~\cite{bochkovskii2024depthpro,yang2024depthanythingv2} struggle to maintain cross-frame consistency, we can utilize the prediction and optimization of DUSt3R to enforce the cross-frame consistency for these monocular depth estimators. 
However, directly applying DUSt3R here presents two main challenges. First, DUSt3R is designed specifically for static scenes, limiting its applicability to dynamic videos. Second, DUSt3R's point map predictions tend to yield less detailed depth maps compared to pixel-aligned monocular depth estimators, reducing accuracy. To address these limitations, we propose to integrate monocular depth estimation within the DUSt3R framework and fine-tune DUSt3R for dynamic scenes.

\begin{figure*}[!ht]
    \begin{center}
        \includegraphics[width=1\textwidth]{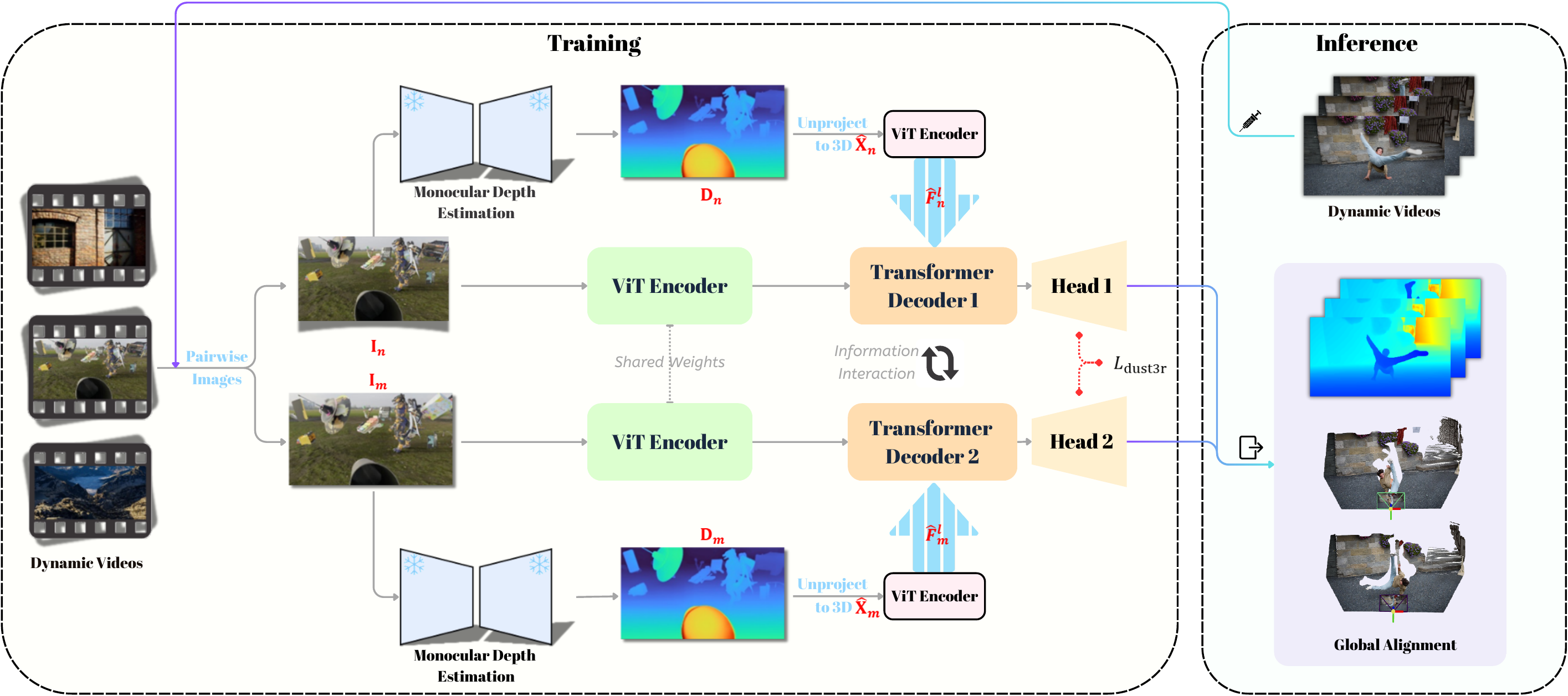}
        \caption{\textbf{Architecture of Align3R.} Given two frames of a video, we apply the ViT-based encoder and decoder to predict pairwise point maps from them. In this process, we apply the external monocular depth estimator to estimate depth maps for these two images, process the estimated depth with a new ViT-based encoder, and finally inject the extracted features from this new encoder into the decoder of the original DUSt3R decoder with zero convolution layers. During inference, we apply global alignment to ensure consistent depth maps, camera poses and point clouds across each frame.
        }
        \label{framework}
    \end{center}
    \vspace{-1.5em}
\end{figure*}

\subsection{Incorporating Monocular Depth Estimation}
In this section, we aim to design a mechanism to incorporate the predicted monocular depth maps from Depth Pro~\cite{bochkovskii2024depthpro} or Depth Anything V2~\cite{yang2024depthanythingv2} into the DUSt3R model. A straightforward approach to incorporating monocular depth maps is to directly concatenate depth maps with the input RGB images. However, as shown in experiments, this approach destructively breaks the feature distributions of the pre-trained DUSt3R encoder. Instead, inspired by ControlNet~\cite{zhang2023controlnet}, we adopt a new vision transformer to extract features from the depth maps and inject these features into the decoder of DUSt3R without ruining the original prediction before fine-tuning, as illustrated in Fig.~\ref{framework}.

\textbf{Depth to points}. Since the predictions of the DUSt3R model are point maps for two images, we transform the estimated depth into point maps of the modality for better convergence. We unproject the estimated depth maps into 3D space to generate 3D point maps $\mathbf{\hat{X}}_n, \mathbf{\hat{X}}_m \in \mathbb{R}^{\rm{H\times W\times 3}}$. This unprojection process requires the intrinsics of the input images. In models with focal length prediction, like Depth Pro~\cite{bochkovskii2024depthpro}, we utilize the predicted focal length to construct the intrinsic matrix. 
Meanwhile, for models without focal length prediction, like Depth Anything V2~\cite{yang2024depthanythingv2}, we set the focal length to a fixed value. 
Due to the large numerical ranges of predicted depth values, We normalize each axis $(x, y, z)$ of the unprojected point maps separately to the range $[-1, 1]$, ensuring stable model training.

\textbf{Point map ViT}. Next, for each point 
map $\mathbf{\hat{X}}_i$ with $i=n$ or $i=m$, we apply a standard patch embedding method to divide the point maps into patches
\begin{equation}
\label{eq:patch}
\mathbf{\hat{X}}_i'=\text{PatchEmbed}(\mathbf{\hat{X}}_i),
\end{equation}
where $\mathbf{\hat{X}}'_i\in\mathbb{R}^{\rm{H'\times W'\times C}}$ is the patchified point map.
Subsequently, we apply the self-attention mechanism to $\mathbf{\hat{X}}'_i$ to generate multi-level features $\mathbf{\hat{F}}_i^{(1)},\mathbf{\hat{F}}_i^{(2)},\dots,\mathbf{\hat{F}}_i^{(s)}$, which will be injected into the DUSt3R decoder for information aggregation. At this stage, to avoid ruining the prediction of the original DUSt3R model, we use zero convolution~\cite{zhang2023adding} for feature fusion
\begin{equation}
\label{eq:inject}
\mathbf{\hat{E}}_i^{(l)}=\text{ZeroConv}(\mathbf{\hat{F}}_i^{(l)})+\mathbf{E}_i^{(l)},l=1,2,\dots,s,
\end{equation}
where $\mathbf{E}_i^{(l)}$ is the feature map generated by the DUSt3R decoder on the $l$-th layer and $\mathbf{\hat{E}}_i^{(l)}$ is the feature map with the injected point map features.

\subsection{Fine-tuning on Dynamic Videos}
The original DUSt3R model is only trained on static scenes and cannot correctly process dynamic videos. To improve the performance, we fine-tune the DUSt3R model along with the additional point map transformer on dynamic videos with ground-truth depth maps. To maintain DUSt3R's robust feature extraction capabilities, the encoder is frozen while only the decoder and the additional point map transformer are fine-tuned. 
The fine-tuning loss following DUSt3R is defined as follows
\begin{equation}
L_{dust3r} = \left\Vert \frac{1}{z}\mathbf{X}^{e}_v  - \frac{1}{\overline{z}}\overline{\mathbf{X}}^{e}_v \right\Vert_2,
\label{eq:regression}
\end{equation}
where \(v \in \{n,m\}\) denotes the view index, \(\mathbf{X}\) and \(\overline{\mathbf{X}}\) are the predicted and ground-truth point maps, and \(z\) and \(\overline{z}\) are scaling factors used to normalize the predicted and ground-truth point maps. Notably, \(z\) and \(\overline{z}\) are computed from all valid pixels within a single image. 

\textbf{Datasets}. We apply 5 synthetic datasets for fine-tuning, including Sceneflow~\cite{mayer2016large}, VKITTI~\cite{gaidon2016virtual}, TartanAir~\cite{wang2020tartanair}, Spring~\cite{mehl2023spring}, PointOdyssey~\cite{zheng2023pointodyssey}, as shown in Tab.~\ref{table:Datasets}. Among all these 5 datasets, we also include a diverse static dataset TartanAir to keep the performances on static regions. Among all scenes in the SceneFlow dataset, we use Monkaa, Driving, and a part of FlyingThings3D for training and we adopt the remaining part of FlyingThings3D as the test set. The large-scale synthetic dataset PointOdyssey contains a training set and a validation set so we adopt the training set for fine-tuning and the validation set in the evaluation. 
On all these datasets, we sample two frames with temporal strides ranging from 1 to 10 as the training image pairs, which leads to reasonable overlap regions between the image pairs.

\textbf{Depth filtering}.
A noticeable problem in fine-tuning DUSt3R is that the numerical ranges of ground-truth point maps are very large. Far-away points with large depth values will dominate the training loss of Eq.~\eqref{eq:regression}. However, for an image pair with a limited baseline length, it is hard to predict points with large depth values exactly due to their small disparities and these far-away points are not as important as nearby objects. Thus, we filter out depth regions beyond 400 meters, which reduces the impact of distant objects (such as the sky) on prediction accuracy. This approach enables our model to better adapt to open-world dynamics by emphasizing depth prediction of objects near to camera.

\subsection{Inference on Long Videos}

After fine-tuning the model on dynamic videos, we apply the model to predict pairwise point maps for the given dynamic video. Then, we follow the DUSt3R to enforce the constraint in Eq.~\eqref{eq:DUSt3R_global} to solve for the consistent depth maps and camera poses for each frame. However, we find that for long videos with more than 30 frames, the original optimization strategy in DUSt3R consumes too much memory and causes out-of-memory on a 4090 GPU. We adopt a hierarchical optimization to reduce memory consumption.

\textbf{Hierarchical optimization}. Given a long video sequence, we divide the video into $K$ clips of a predefined length of $M=10$ or $M=20$. For each clip, we take one image as the key frame and construct a keyframe clip of length $K$. By conducting global alignment on the keyframe clip, we initialize the depth maps, camera poses, and focal lengths of these keyframes. Then, for each divided clip, we apply a local alignment to compute the depth maps, camera poses, and focal lengths for the remaining frames. 
Such a global-local hierarchical optimization ensures that we only optimize a short clip with limited frames every time, which effectively reduces memory and time consumption while keeping consistency.

\begin{table}
\footnotesize
\setlength\tabcolsep{5pt}
\caption{\textbf{Statistics of datasets for fine-tuning.}}
\label{table:Datasets}
\centering
    \begin{tabular}{lcccc}
    \toprule
      \multirow{2}{*}{Name} & \multirow{2}{*}{\#Scene}& Avg. & Total  &\multirow{2}{*}{Type} \\
                            &                        & \#Img& \#Img  & \\
      \midrule
    SceneFlow~\cite{mayer2016large} &8k&10&80k&Dynamic\\
    VKITTI~\cite{gaidon2016virtual} &100&425&42k&Dynamic\\
    TartanAir~\cite{wang2020tartanair} &500&900&450k&Static\\
    Spring~\cite{mehl2023spring} &37&135&5k&Dynamic\\
    PointOdyssey~\cite{zheng2023pointodyssey} &131&1.8k&240k&Dynamic\\
      \bottomrule
  \end{tabular}

\end{table}

%% file: sec/4_experiments.tex
\section{Experiments}
\subsection{Implementation details}
We train our model on six RTX 4090 GPUs with a batch size of 12, requiring approximately 20 hours for 50 epochs. We use the AdamW optimizer with a learning rate of 0.00005. Input images are resized randomly to one of three resolutions: $512 \times 288$, $512 \times 336$, or $512 \times 256$. Each epoch consists of 27,750 sampled image pairs, which contain 5,000 pairs from SceneFlow, 3,000 from VKITTI, 6,250 from TartanAir, 1,000 from Spring, and 12,500 from PointOdyssey. We use $s = 6$ layers of features in Eq.~\eqref{eq:inject}. We adopt the flow losses proposed in MonST3R~\cite{zhang2024monst3r} to introduce the RAFT~\cite{teed2020raft} flow in inference optimization, which influences the depth quality a little but is important for accurate camera pose estimation. More implementation details can be found in the supplementary material.

\subsection{Video depth estimation}
\textbf{Evaluation datasets.} We evaluate our model on both real-world datasets, including the Bonn~\cite{palazzolo2019refusion} and TUM dynamics~\cite{sturm2012benchmark} datasets, and synthetic datasets, including the Sintel~\cite{butler2012naturalistic} dataset, the PointOdyssey~\cite{zheng2023pointodyssey} validation set, and the test set FlyingThings3D of Sceneflow~\cite{mayer2016large}. Additionally, we report the qualitative results on the DAVIS~\cite{perazzi2016benchmark} dataset.

The \textbf{Bonn} dataset is a real-world RGB-D SLAM dataset containing 24 dynamic sequences, where people are doing different tasks such as manipulating boxes or playing with balloons. We evaluate our model on five videos with an average of 110 frames per video, the same as the setting used in DepthCrafter~\cite{hu2024depthcrafter}. 
The \textbf{TUM dynamics} dataset is also a real-world dataset with 8 dynamic scenes and we select 50 frames from each scene for evaluation.
The \textbf{Sintel} dataset is a synthetic dynamic dataset with 23 videos and approximately 50 frames per video.
The \textbf{PointOdyssey} validation set is a synthetic dataset with 15 dynamic scenes with numerous moving foreground objects. We evaluate our model on the first 110 frames of each video. 
The \textbf{FlyingThings3D} test set is also a synthetic dataset with many moving foreground objects, containing around 900 dynamic scenes with approximately 10 frames per scene. We select 44 scenes with a stride of 20 for evaluation.

\begin{table*}[!t]
  \begin{center}
    \footnotesize
    \setlength\tabcolsep{1.5pt}
    \caption{\textbf{Video depth estimation results.} We evaluate our model on both real-world datasets, Bonn and TUM dynamics, and synthetic datasets, Sintel, PointOdyssey validation set, and Sceneflow test set. \textbf{Best} and \underline{second best} results are highlighted.}
    \label{table:depth}
    \scalebox{0.9}{\begin{tabular}{ll|cc|cc|cc|cc|cc}
      \toprule
       \multirow{3}{*}{Category} &\multirow{3}{*}{Method} & \multicolumn{6}{c|}{Indoors \& outdoors (Hard)}& \multicolumn{4}{c}{Indoors  (Easy)}\\
       \cline{3-12}
       &&\multicolumn{2}{c|}{Sintel} &   \multicolumn{2}{c|}{PointOdyssey val} &  \multicolumn{2}{c|}{FlyingThings3D test} &\multicolumn{2}{c|}{Bonn 5 scenes} &  \multicolumn{2}{c}{TUM dynamics}  \\

       & & Abs Rel $\downarrow$& $\delta<1.25 \uparrow$   & Abs Rel $\downarrow$& $\delta<1.25 \uparrow$& Abs Rel $\downarrow$& $\delta<1.25 \uparrow$& Abs Rel $\downarrow$& $\delta<1.25 \uparrow$& Abs Rel $\downarrow$& $\delta<1.25 \uparrow$ \\
       \midrule
       \multirow{2}{*}{Single-frame depth} &Depth Anything V2~\cite{yang2024depthanythingv2}&0.348&0.592&0.214&0.688&0.267&0.616&0.118&0.882&0.184&0.750\\
       &Depth Pro~\cite{bochkovskii2024depthpro}&0.418&0.559&0.167&0.779&0.322&0.537&\textbf{0.067}&\textbf{0.974}&\textbf{0.106}&\underline{0.887}\\
       \midrule
       \multirow{2}{*}{Video depth} &ChronoDepth~\cite{shao2024chronodepth}& 0.687 &0.486 &0.210&0.707&0.288&0.633&0.100 &0.911&0.151&0.825
\\
       &DepthCrafter~\cite{hu2024depthcrafter}&0.292 &\textbf{0.697} &0.229&0.675&/&/&0.075 &0.971&0.176
&0.744\\
       \midrule
       &DUSt3R~\cite{wang2024dust3r}&0.422&0.542&0.184&0.743&0.140&0.817&0.154&0.839&0.202&0.775\\
        Joint video depth &MonST3R~\cite{zhang2024monst3r}&0.335&0.586 &0.089&0.909&0.132&0.836&0.082 &0.953 &0.140&0.841\\
        depth \& pose &Ours (Depth Anything V2)&\textbf{0.253}&\underline{0.681}&\underline{0.078}&\underline{0.929}&\underline{0.106}&\underline{0.890}&0.075&\underline{0.972}&\underline{0.109}&\textbf{0.915}\\
        &Ours (Depth Pro)&\underline{0.263}&0.641&\textbf{0.077}&\textbf{0.930}&\textbf{0.102}&\textbf{0.895}&\underline{0.068}&0.969&0.112&0.884\\
      \bottomrule
    \end{tabular}}
     \vspace{-1.2em}
  \end{center}
\end{table*}
\begin{table*}[!t]
  \begin{center}
    \footnotesize
    \setlength\tabcolsep{2.5pt}
    \caption{\textbf{Camera pose estimation results.} We evaluate our model on two real-world datasets: TUM dynamics and Bonn, as well as the synthetic Sintel dataset. \textbf{Best} and
\underline{second best} results are highlighted. $\ddagger$ means using ground truth camera intrinsics as input. 
}
    \label{pose_estimation}

    \begin{tabular}{ll|ccc|ccc|ccc}
      \toprule
       \multirow{2}{*}{Category} &\multirow{2}{*}{Method} &  \multicolumn{3}{c|}{TUM dynamics} &\multicolumn{3}{c|}{Bonn 5 scenes}&  \multicolumn{3}{c}{Sintel}  \\

       & & ATE $\downarrow$& RTE$\downarrow$   & RRE$\downarrow$& ATE($10^{-2}$) $\downarrow$&RTE($10^{-2}$)$\downarrow$   & RRE$\downarrow$ & ATE $\downarrow$&RTE$\downarrow$   & RRE$\downarrow$\\
       \midrule
       
     &  DROID-SLAM$\ddagger$~\cite{teed2021droid} &/&/&/ &/&/&/&0.175 &0.084& 1.912\\
    Pose only&   DPVO$\ddagger$~\cite{teed2024deep} &/&/&/ &/&/&/&0.115 &0.072 &1.975\\
    & COLMAP~\cite{schoenberger2016sfm} & 0.076&0.059&7.689  &3.43&0.927&0.905&0.559&0.325&7.302\\
    \midrule
     &Robust-CVD~\cite{kopf2021robust}& 0.153& 0.026 &3.528 &/&/&/&0.360 &0.154 &3.443\\
    &CasualSAM~\cite{zhang2022structure} &0.071& 0.010& 1.712 &/&/&/&\underline{0.141}& \textbf{0.035}& 0.615 \\
    Joint depth&DUSt3R~\cite{wang2024dust3r}&0.093	&0.035&1.708&2.166&	0.650&	1.169&0.601&	0.214	&11.426\\
    \& pose&MonST3R~\cite{zhang2024monst3r}&0.020&0.014&0.478&0.686	&0.595	&0.593&\textbf{0.111}&0.044&0.780\\
    &Ours (Depth Anything V2)&\textbf{0.011}&\underline{0.010}&\textbf{0.321}&\textbf{0.646}&	\underline{0.588}&	\underline{0.585}&0.163&0.062&\textbf{0.419}\\
    &Ours (Depth Pro)&\underline{0.012}&\textbf{0.010}&\underline{0.327}&\underline{0.673}	&\textbf{0.570}	&\textbf{0.576}&0.128&\underline{0.042}&\underline{0.432}\\
      \bottomrule
    \end{tabular}
     \vspace{-1.2em}
  \end{center}
\end{table*}

\textbf{Evaluation metrics.} Following previous methods~\cite{hu2024depthcrafter,zhang2024monst3r}, we evaluate our model by aligning the estimated depth maps with the ground truth using a single scale and shift before calculating the metrics. To ensure a fair comparison of temporal consistency across methods, we calculate a shared scale and shift for the entire video sequence, rather than computing individual scale and shift values for each frame as done in monocular depth estimation. We mainly report two metrics, i.e. Abs Rel $\downarrow$(absolute relative error) and percentage of inlier points $\delta<1.25 \uparrow$. 

\textbf{Baselines}. We compare our Align3R with single-frame and video depth estimation methods, i.e. Depth Anything V2~\cite{yang2024depthanythingv2}, Depth Pro~\cite{bochkovskii2024depthpro}, ChronoDepth~\cite{shao2024chronodepth} and DepthCrafter~\cite{hu2024depthcrafter}. We also compare with methods for joint video depth and pose estimation methods, i.e. DUSt3R~\cite{wang2024dust3r} and MonST3R~\cite{zhang2024monst3r}.
For all baseline methods, we adopt their official implementation and re-evaluate their performance using a per-sequence scale and shift alignment on our evaluation datasets.

\begin{figure*}[!t]
    \begin{center}
        \includegraphics[width=1\textwidth]{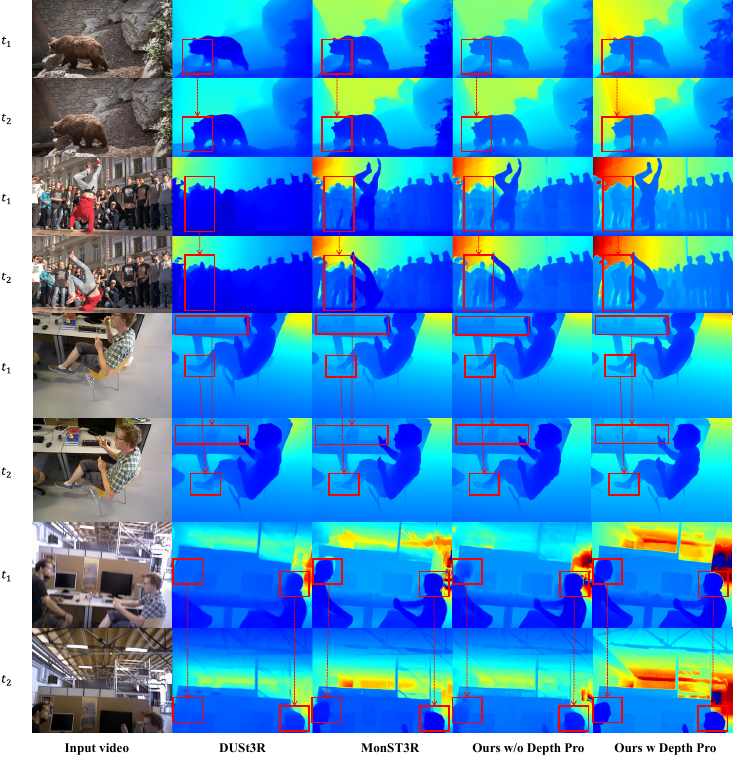}
        \caption{\textbf{Qualitative comparison on the DAVIS and TUM dynamics dataset.} The red boxes show highlighted regions.}
        \label{fig:depth}
    \end{center}
    \vspace{-1.5em}
\end{figure*}

\textbf{Comparison with existing methods.} 
In Tab.~\ref{table:depth}, we compare our results using Depth Anything V2 and Depth Pro as monocular depth input with all baseline methods. We summarize the results of Tab.~\ref{table:depth} in the following.
\begin{enumerate}
    \item Our method achieves superior video depth estimation results compared to previous joint video depth and pose estimation methods DUSt3R and the concurrent work MonST3R on all metrics of all datasets. 
    \item Furthermore, with the assistance of global alignment, we can effectively align the predictions of monocular depth estimation models like Depth Anything V2 and Depth Pro, achieving better temporal consistency on three challenging outdoor datasets, including Sintel, PointOdyssey, and FlyingThings3D, which contain significant dynamic object motions or camera motions. 
    \item On the relatively simpler scenarios, like the Bonn and TUM dynamics datasets, which are indoor scenes with smooth camera motions and fewer dynamic objects, the original predictions of Depth Pro are already high-quality and consistent due to the relatively small and slow motions. We find that our method achieves comparable results as Depth Pro and may lose some details in the global alignment.
\end{enumerate}

We visualize the qualitative comparison on the DAVIS and TUM dynamics datasets in Fig.~\ref{fig:depth}. As shown by the figure, our method achieves more consistent video depth than other baseline methods. Additionally, by incorporating monocular depth, our approach reconstructs more details on the depth maps than MonST3R. 

\subsection{Camera pose estimation}
\textbf{Evaluation datasets.}
We evaluate the camera pose estimation on two real-world datasets, i.e. TUM dynamics and Bonn, and a synthetic dataset, i.e. Sintel. The TUM dynamics dataset consists of 8 sequences. We evaluate all methods on 30 frames of each sequence. For the Bonn dataset, we evaluate our model on five videos and also select 30 frames from each scene. For the Sintel dataset, we follow the evaluation protocol of previous methods~\cite{chen2024leap,zhao2022particlesfm} to exclude scenes that are static or only contain straight motions, resulting in 14 sequences for evaluation.

\textbf{Evaluation metrics.} We follow previous works~\cite{chen2024leap,zhao2022particlesfm,teed2024deep} to report three metrics, i.e. ATE $\downarrow$ (absolute translation error), RTE$ \downarrow$ (relative translation error) and RRE $\downarrow$ (relative rotation error). ATE computes the deviation of estimated trajectories from the ground truth trajectories after alignment.
RPE and RPE are the averaged local translation and rotation errors respectively over consecutive poses.

\textbf{Baselines}. 
We compare our method with both pure camera pose estimation methods including DROID-SLAM~\cite{teed2021droid}, DPVO~\cite{teed2024deep}, COLMAP~\cite{schoenberger2016sfm} and also joint depth and pose estimators including Robust-CVD~\cite{kopf2021robust}, CasualSAM~\cite{zhang2022structure}, DUSt3R~\cite{wang2024dust3r}, and MonST3R~\cite{zhang2024monst3r}.
For DROID-SLAM, DPVO, Robust-CVD, and CasualSAM, we use the results reported in MonST3R while re-evaluating COLMAP, DUSt3R, and MonST3R using their official codes.

\begin{table*}[!tbp]
\setlength\tabcolsep{2.5pt}
    \footnotesize
\caption{\textbf{Ablation study on the Sintel and TUM dynamics dataset.} ``F.t. all" means fine-tuning the whole model of DUSt3R. ``F.t. last 4 layers" means fine-tuning the last 4 layers of the decoder while ``F.t. decoder" means fine-tuning the whole decoder. ``w/o depth" means discarding the estimated depth in the model. ``Concat depth" means concatenating the depth with the input RGB images as inputs to the DUSt3R model. ``ViT encoder" means converting the depth maps into point maps, applying the new ViT to process the point maps, and finally injecting the extracted features into the decoder with zero convolution layers.}%
  \centering
    \label{table:ablation}
  \begin{tabular}{l|cc|cc|ccc|ccc}
      \toprule
      \multirow{3}{*}{Setting} &\multicolumn{4}{c|}{Depth estimation} &\multicolumn{6}{c}{Pose estimation}\\ 
    &\multicolumn{2}{c|}{Sintel}&\multicolumn{2}{c|}{TUM dynamics}&\multicolumn{3}{c|}{Sintel}&\multicolumn{3}{c}{TUM dynamics}\\
    & Abs Rel$\downarrow$ & $\delta$$<$1.25$\uparrow$ & Abs Rel$\downarrow$& $\delta$$<$1.25$\uparrow$   & ATE$\downarrow$& RPE Trans$\downarrow$   & RPE Rot$\downarrow$& ATE$\downarrow$& RPE Trans$\downarrow$   & RPE Rot$\downarrow$\\
    \midrule
    F.t. all  &0.310&0.619&0.153
&0.811&0.257&0.098&	0.801&0.025&	0.020	&0.757\\
    F.t. last 4 layers &0.319
&0.603&0.159
&0.805&0.191&0.086&1.293&0.016&0.016&	0.514\\
    F.t. decoder&0.306&0.613&0.135
&0.864
&0.227&0.117&0.865&0.017&	0.012&	0.435\\
    \midrule
    w/o depth &0.306&0.613&0.135&0.864&0.227&0.117&0.865&0.017&	0.012&	0.435\\
    Concat depth &0.399&0.537&0.182&0.794&0.338&0.246&1.823&0.034&0.025&0.776
\\
    
    ViT encoder & \textbf{0.263} &\textbf{0.641} &\textbf{0.112}
&\textbf{0.884}
&\textbf{0.128}	&\textbf{0.042}	&\textbf{0.432}&\textbf{0.012}&\textbf{0.010}&	\textbf{0.327}\\
    
      \bottomrule
  \end{tabular}
\end{table*}
\textbf{Comparison with existing methods.} Tab.~\ref{pose_estimation} shows the quantitative results of our method and all baselines. The results show that the proposed Align3R achieves consistently better performance than all baseline methods on RTE and RRE metrics. For the ATE, our method achieves the best results on the real-world TUM Dynamics and Bonn datasets but is slightly worse than MonST3R on the synthetic Sintel dataset.
More qualitative comparisons of camera pose estimation are reported in the supplementary material.

\subsection{Ablation study}
We conduct analyses on fine-tuning strategy, how to incorporate monocular depth maps, and memory reduction of our hierarchical optimization strategy.
More analyses are provided in the supplementary material.

\textbf{Finetuning strategy}. 
In the top part of Tab.~\ref{table:ablation}, we show the results of three different fine-tuning settings. From the results, it can be seen that fine-tuning the full model results in lower performance because such full fine-tuning disrupts the encoder features from DUSt3R. Then, we observe that fine-tuning all decoder layers leads to better performance than fine-tuning only a subset of decoder layers because fine-tuning the full decoder enhances the ability of the model to adapt to dynamic videos. 

\begin{figure}[!t]
    \begin{center}
        \includegraphics[width=1\textwidth]{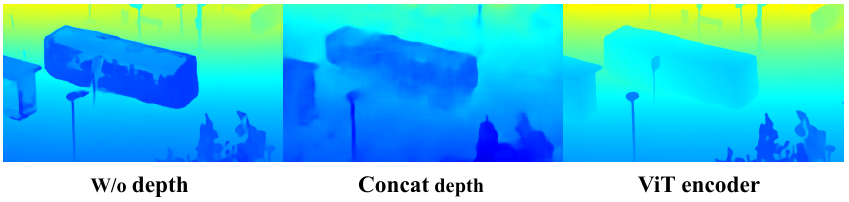}
        \caption{\textbf{Qualitative ablation study on estimated depth.} }
        \label{fig:ablation}
    \end{center}
    \vspace{-1.5em}
\end{figure}

\textbf{Depth incorporation}. 
We conduct an ablation study on whether to incorporate the monocular depth and how to incorporate it. The results are shown in the bottom part of Tab.~\ref{table:ablation}. As shown by the results, discarding the input monocular depth degenerates the quality of depth maps because such estimated monocular depth maps often contain more details and thus help predict details on the point maps.
Then, we compare an alternative strategy to concatenate the depth map with the input RGB images. Directly concatenating the depth maps disrupts the distribution of the pre-trained DUSt3R encoder, leading to a degenerated performance. In contrast, the proposed strategy extracts features from the depth maps with a transformer and injects the features into the decoder of DUSt3R, which significantly improves the results. A qualitative comparison between these three settings is shown in Fig.~\ref{fig:ablation}.

\textbf{Hierarchical optimization}. To validate our design of hierarchical optimization, we conduct an experiment on the Bonn dataset by selecting 30 frames for each video. In our hierarchical optimization, we set the predefined clip length $M = 10$. The results in Tab.~\ref{table:timememory} show that in comparison with the original DUSt3R optimization strategy, our hierarchical optimization strategy effectively reduces memory and time consumption while maintaining quality.

\begin{table}[!tbp]
\footnotesize
\setlength\tabcolsep{1.5pt}
\caption{\textbf{Ablation study on hierarchical optimization (``HO").}}%
  \centering
    \label{table:timememory}
  \begin{tabular}{c|cccc}
      \toprule
     Setting &Abs Rel$\downarrow$ & $\delta$$<$1.25$\uparrow$ & Avg. time (min)$\downarrow$& Memory (GB)$\downarrow$\\
    \midrule
    w/o HO&\textbf{0.054}&\textbf{0.975}&2.9 & 24.0 \\
    w. HO&0.056&0.974&\textbf{1.1}&\textbf{5.9} \\
      \bottomrule
  \end{tabular}
\end{table}

%% file: sec/5_conclusion.tex
\section{Conclusion}
In this paper, we introduce a new method called \methodname to simultaneously estimate depth maps and camera poses of dynamic videos. The key idea of \methodname is to combine a monocular depth estimator with the DUSt3R model. We propose a novel strategy to apply a transformer to extract features from the monocular depth and inject the extracted features of monocular depth into the decoder of the DUSt3R model. Then, we finetune the DUSt3R model on the dynamic videos. Finally, we can predict pairwise point maps and utilize these point maps to effectively solve for consistent video depth and camera poses. Extensive experiments on 6 synthetic and real-world datasets demonstrate the superior performances of our method.

%% file: sec/X_suppl.tex
\clearpage
\setcounter{page}{1}
\maketitlesupplementary

\section{More implementation details}
To estimate monocular depth, we employ Depth Anything V2~\cite{yang2024depthanythingv2} and Depth Pro~\cite{bochkovskii2024depthpro}. For Depth Anything V2, we use the large model variant to predict depth maps. 
During the global alignment of our method, we perform 300 iterations with the Adam optimizer, setting an initial learning rate of 0.05 and using a cosine learning rate schedule.

\section{More qualitative results}
\subsection{Depth comparison}
\begin{figure*}[!ht]
    \begin{center}
        \includegraphics[width=0.9\textwidth]{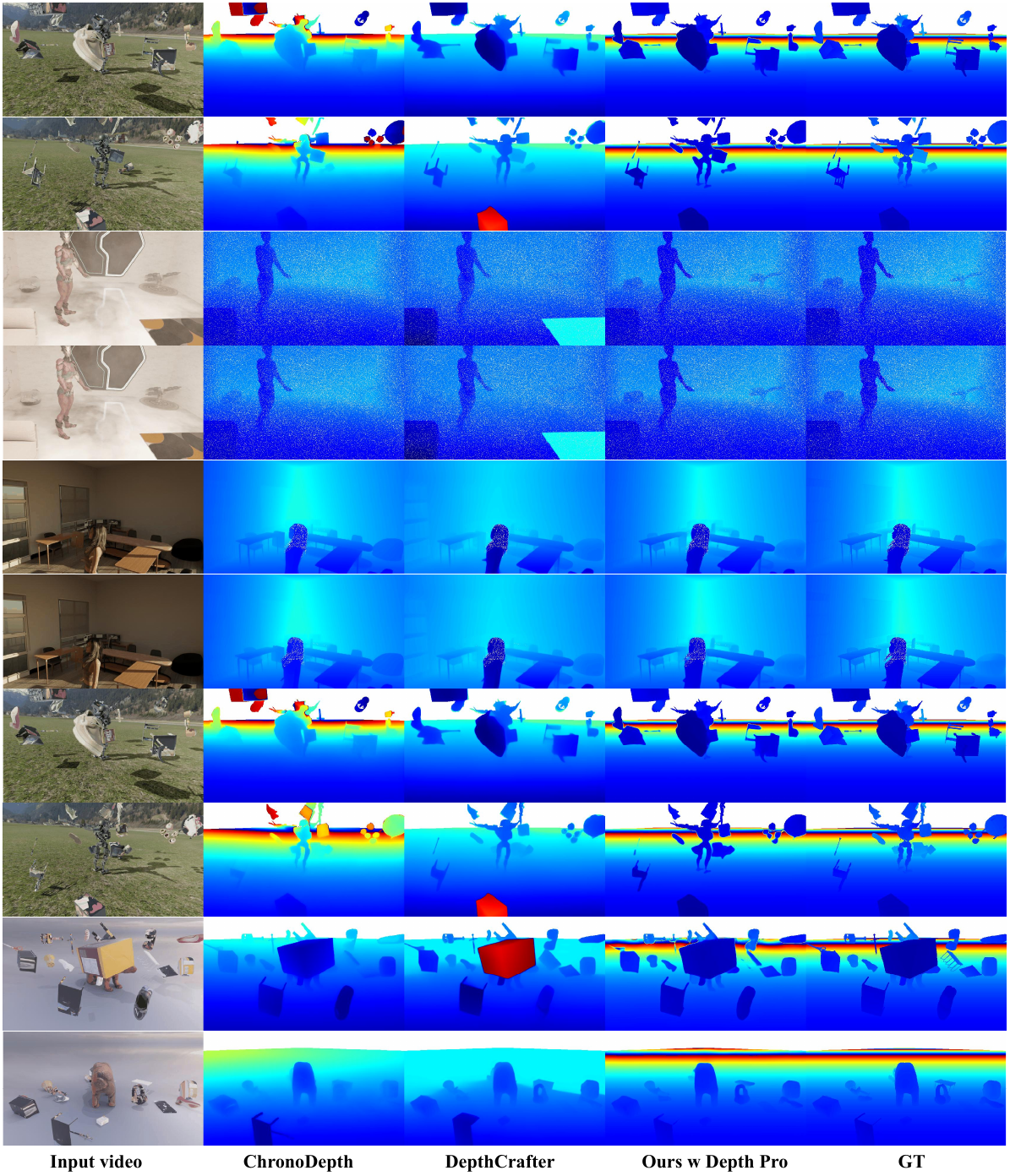}
        \caption{\textbf{Qualitative comparison on the PointOdyssey~\cite{zheng2023pointodyssey} validation set with ChronoDepth~\cite{shao2024chronodepth} and DepthCrafter~\cite{hu2024depthcrafter}.} }
        \label{figure:depth_compare_point}
    \end{center}
    \vspace{-1.5em}
\end{figure*}

\begin{figure*}[!ht]
    \begin{center}
        \includegraphics[width=0.9\textwidth]{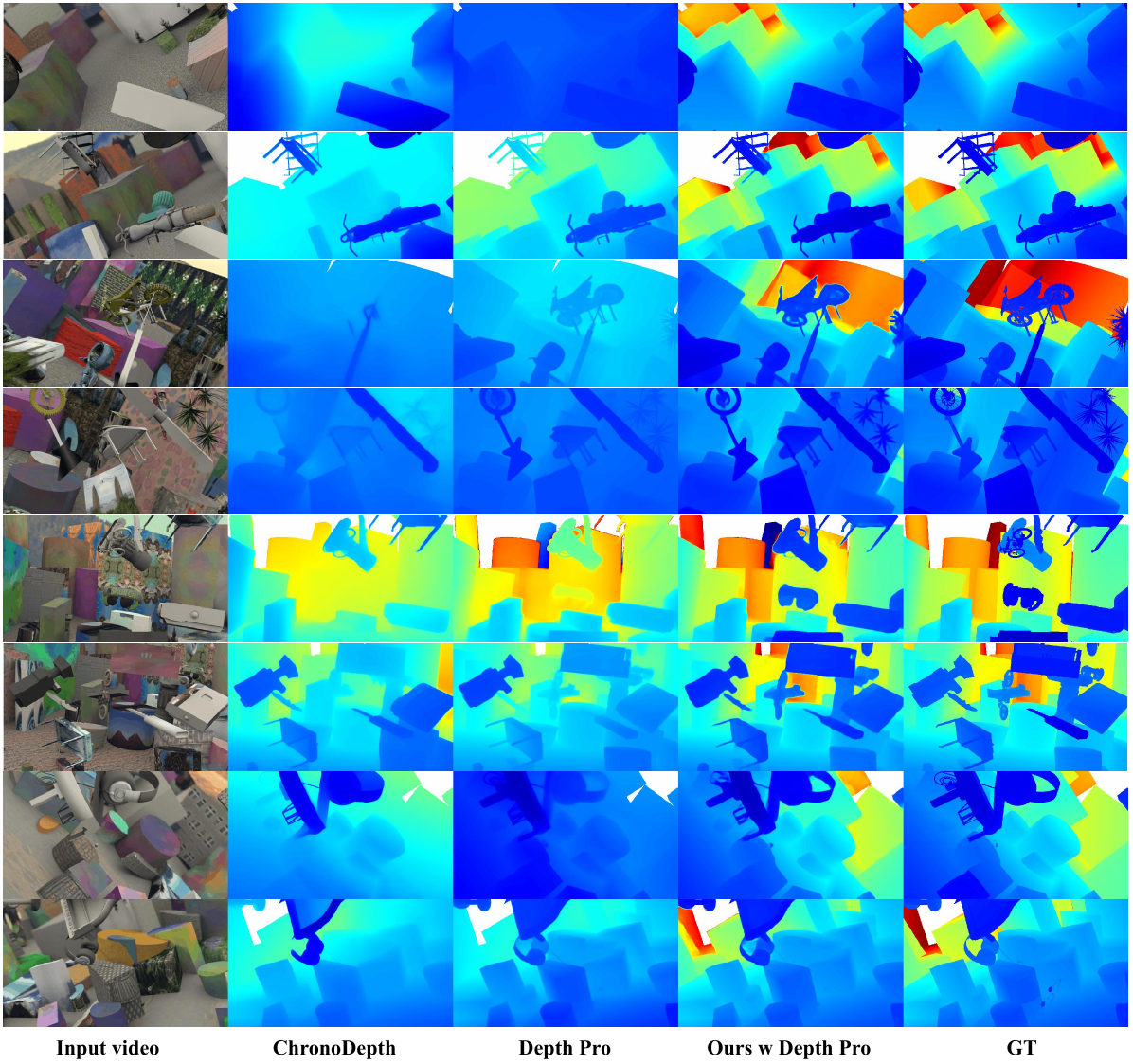}
        \caption{\textbf{Qualitative comparison on the FlyingThings3D~\cite{mayer2016large} test set with ChronoDepth~\cite{shao2024chronodepth} and Depth Pro ~\cite{bochkovskii2024depthpro}.} }
        \label{figure:depth_comparesceneflow}
    \end{center}
    \vspace{-1.5em}
\end{figure*}

To provide a more vivid illustration, we perform visual comparisons on the PointOdyssey~\cite{zheng2023pointodyssey} validation set and the FlyingThings3D~\cite{mayer2016large} test set, both containing numerous moving objects. In Fig.~\ref{figure:depth_compare_point} and Fig.~\ref{figure:depth_comparesceneflow}, we compare the Depth Pro version Align3R with two video depth estimation methods, ChronoDepth~\cite{shao2024chronodepth} and DepthCrafter~\cite{hu2024depthcrafter}. It is worth noting that we visualize the depth after sequence alignment, with invalid areas replaced by white. These comparisons demonstrate that, after alignment, our approach achieves enhanced temporal consistency and finer detail by integrating the monocular depth estimator Depth Pro with DUSt3R~\cite{wang2024dust3r}. Additionally, in Fig.~\ref{figure:depth_compare_point}, the reason why some foreground objects predicted by DepthCrafter are shown in red is primarily due to certain regions in DepthCrafter having depth values less than 0 after sequence alignment. This indicates that the relative depth relationships between objects generated by DepthCrafter are not entirely accurate.

\subsection{Camera pose comparison}
\begin{figure*}[!t]
    \begin{center}
        \includegraphics[width=1\textwidth]{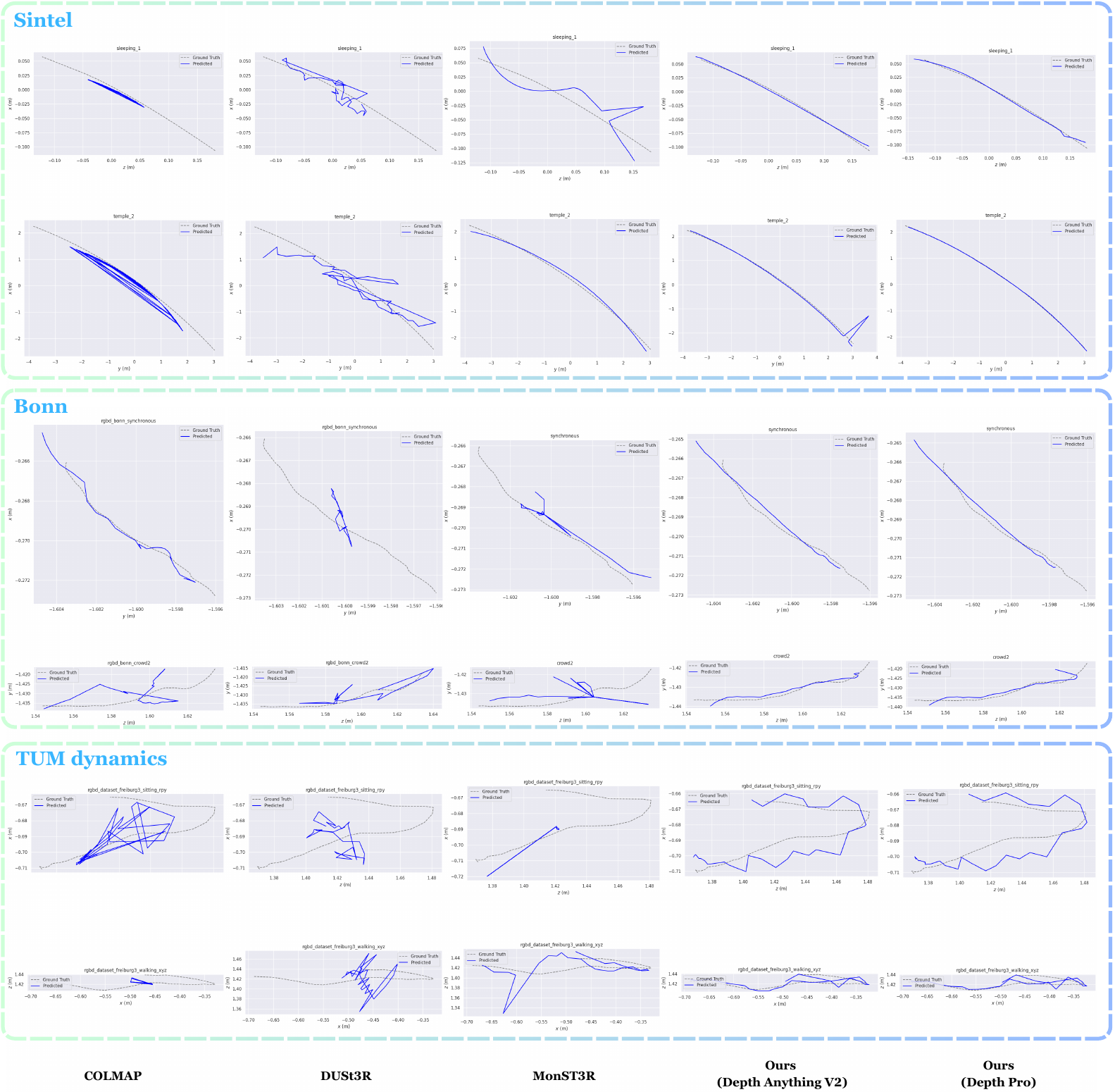}
        \caption{\textbf{Camera pose estimation comparison} on the TUM dynamics~\cite{sturm2012benchmark}, Bonn~\cite{palazzolo2019refusion}, and Sintel~\cite{butler2012naturalistic} datasets.}
        \label{figure:pose_curve}
    \end{center}
\end{figure*}

In Fig.~\ref{figure:pose_curve}, we present qualitative results for camera pose estimation on the Sintel~\cite{butler2012naturalistic}, Bonn~\cite{palazzolo2019refusion}, and TUM dynamics~\cite{sturm2012benchmark} datasets. We compare our model with the pose-only method COLMAP~\cite{schoenberger2016sfm} and two joint depth and pose estimation methods, DUSt3R~\cite{wang2024dust3r} and MonST3R~\cite{zhang2024monst3r}. These comparisons show that our approach achieves improved camera pose estimation, demonstrating better consistency and closer alignment with the ground truth trajectory.

\subsection{Dynamic point clouds}
\begin{figure*}[!ht]
    \begin{center}
        \includegraphics[width=0.9\textwidth]{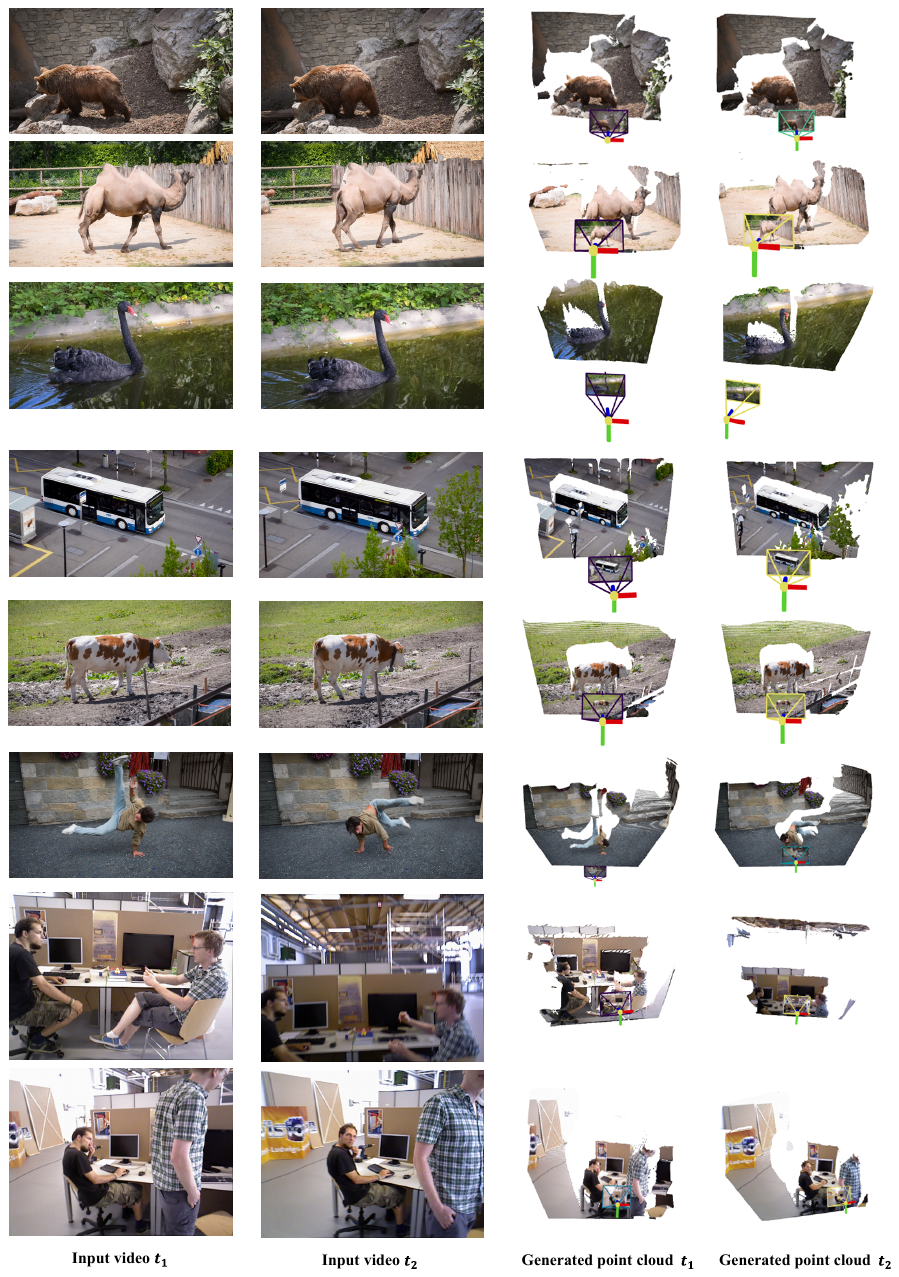}
        \caption{\textbf{Visualization of point clouds on the DAVIS~\cite{perazzi2016benchmark} and TUM dynamics~\cite{palazzolo2019refusion} datasets.} }
        \label{figure:supp_pc}
    \end{center}
    \vspace{-1.5em}
\end{figure*}

To further demonstrate the effectiveness of our method in depth and camera pose estimation, we present additional visualizations of the reconstructed point clouds. As illustrated in Fig.~\ref{figure:supp_pc}, the reconstructed point clouds exhibit strong geometric accuracy and temporal consistency, maintaining a clear structure for dynamic objects. This consistency across frames highlights our model’s ability to handle complex, real-world movements while preserving coherent geometry. Such results underline the robustness of our approach, effectively capturing and maintaining precise depth and pose information for improved 3D scene understanding in dynamic environments.
\begin{table}[!tbp] 
\caption{\textbf{Analysis of the scale map optimization on the Sintel dataset}.}%
  \centering
    \label{table:monocularoptimization}
  \scalebox{0.8}{\begin{tabular}{c|cc}
      \toprule
      \multirow{2}{*}{Optimization} &\multicolumn{2}{c}{Depth estimation} \\ 
    & Abs Rel $\downarrow$& $\delta<1.25 \uparrow$   \\
    \midrule
    Depth maps &0.306 &0.613 \\
    Scale maps &0.419&0.604 \\
      \bottomrule
  \end{tabular}}
\end{table}
\begin{figure*}[!t]
    \begin{center}
        \includegraphics[width=1\textwidth]{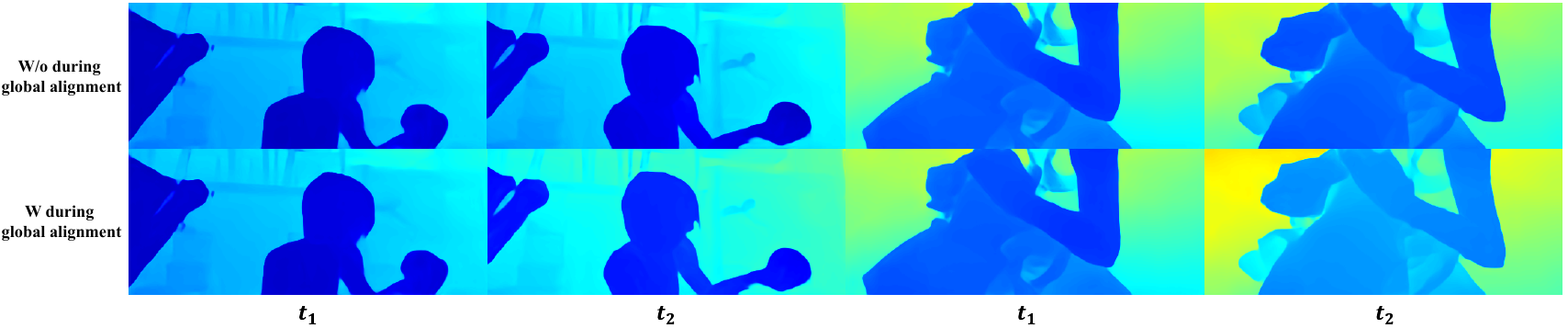}
        \caption{\textbf{Visualization results with and without incorporating monocular depth estimation during global alignment.}}
        \label{figure:monocularoptimization}
    \end{center}
    \vspace{-1.5em}
\end{figure*}
\section{More ablation study}
\textbf{Directly aligning the depth.} 
An alternative to get consistent depth maps is to align the monocular depth map with scale factors.
In Tab.~\ref{table:monocularoptimization}, we align the monocular depth map $I_v$ predicted by Depth Pro using a scale map. Instead of learning a depth map for each frame, we learn $\mathbf{S}:=\{\mathbf{S}_v\in \mathbb{R}^{H\times W}|v=1,...,N\}$ a set of scale maps to minimize the DUSt3R target,
\begin{eqnarray}
     \argmin_{\mathbf{S},\mathbf{\pi},\sigma} \sum_{e \in \mathcal{E}} \sum_{v \in e} \mathbf{C}_v^{e} \left\Vert \mathbf{S}_v \mathbf{\hat{D}}_v - \sigma_e P_e(\pi_v, \mathbf{X}_v^{e}) \right\Vert_2^2.
\label{eq:pose_optim}
\end{eqnarray}
The only difference here is that we do not learn a set of depth maps $\mathbf{D}$ but we learn the scale map $\mathbf{S}_v$ and compute the depth map as the product $\mathbf{D}_v = \mathbf{S}_v \mathbf{\hat{D}}_v$ where $\mathbf{\hat{D}}_v$ is the predicted Depth Pro depth map on the $v$-th view. This optimization process corresponds to traditional video depth optimization methods~\cite{luo2020consistent,kopf2021robust}.
However, since the initialized monocular depth maps predicted by Depth Pro are inconsistent across different frames, As shown in Fig.~\ref{figure:monocularoptimization}, solely optimizing the scale maps leads to inferior performances.

\textbf{Flow loss of MonST3R~\cite{zhang2024monst3r}.}
We adopt the flow loss in MonST3R~\cite{zhang2024monst3r} because we find that flow loss does not affect the depth estimation too much but plays a crucial role in achieving accurate camera pose estimation. As shown in Table~\ref{table:flow}, we conduct an experiment on the Sintel dataset to analyze the effects of flow loss. Since camera poses can only be evaluated in 14 scenes (as discussed in Section 4.3 of the main text), we also report the depth results of these same 14 scenes. From the comparison, we observe minimal differences in depth metrics but significant improvements in pose estimation. Meanwhile, we find that directly applying the flow loss to the original DUSt3R greatly improve the pose estimation. The main reason is that the camera poses can be determined by several robust correspondences while being insensitive to the most depth values.
\begin{table*}[!htbp] 
\caption{\textbf{Analysis of the flow loss~\cite{zhang2024monst3r} for depth and pose estimation.}}%
  \centering
    \label{table:flow}
  \scalebox{1}{\begin{tabular}{c|cc|ccc}
      \toprule
      Setting & Abs Rel$\downarrow$& $\delta$$<$1.25$\uparrow$   & ATE$\downarrow$& RPE Trans$\downarrow$   & RPE Rot$\downarrow$ \\
    \midrule
   DUSt3R w/o flow&0.515&0.533&0.601&	0.214	&11.426\\
   DUSt3R w flow&0.512&0.549&0.327&0.111&1.014\\
   MonST3R &0.353&0.570&\textbf{0.111}&0.044&0.780\\
   Ours w/o flow&\textbf{0.314}&0.562&0.204&0.164& 2.305\\
   Ours w flow&0.317&\textbf{0.577}&0.128&\textbf{0.042}&\textbf{0.432}\\

      \bottomrule
  \end{tabular}}
\end{table*}

\textbf{Runtime analysis.} 
In Tab.~\ref{table:inferencetime}, we provide an additional comparison of inference time using the same dataset setting as Tab.5 in the main text. Since the number of image pairs is a primary factor influencing inference time, we count the image pairs for each method to better understand the reasons behind these differences. In DUSt3r, with a window size of 10, for any given image \( i \), the image pairs are: 
\begin{equation}
\begin{aligned}
\{(i, (i+1) \% 30), ((i+1) \% 30, i), \dots, (i, (i+10) \% 30), \\((i+10) \% 30, i)\},
\end{aligned}
\end{equation}
resulting in a total of \(10 \times 30 \times 2 = 600\) pairs.
In MonST3r, the stride is set to 2 with a window size of 5, and explicit loop closure is not considered. So for any image \( i \), the pairs are: 
\begin{equation}
\begin{aligned} \{(i, i+1), (i+1, i), (i, i+1+2), (i+1+2, i), \dots, \\(i, i+2k+1), (i+2k+1, i)\},
\end{aligned}
\end{equation} 
where \(k = \min(5, \frac{30-1-i}{2})\). This configuration yields a total of 250 image pairs.
In our method, we divide the 30 images into 3 groups, without explicit loop closure and symmetrical pairs. So the total image pairs are \(3\) (keyframe pairs) \( + \frac{10 \times 9}{2} \times 3 = 138\) (each group pairs) pairs. 
Thus, due to the significant difference in the number of image pairs, our method achieves the fastest inference speed, regardless of whether flow and trajectory smoothness losses are applied.

\begin{table}[!htbp] 
\caption{\textbf{Comparison on inference time.} 
}%
  \centering
    \label{table:inferencetime}
  \scalebox{0.8}{\begin{tabular}{l|cc}
      \toprule
      Method & \#Pair &  Avg. time (min)$\downarrow$   \\
    \midrule
    
    DUSt3R~\cite{wang2024dust3r} &600&2.9\\
    MonST3R~\cite{zhang2024monst3r} &250&2.6\\
    Ours&138& 1.8\\
      \bottomrule
  \end{tabular}}
\end{table}

\section{Relationship with MonST3R}
Align3R is a concurrent work with MonST3R~\cite{zhang2024monst3r}. We started our project in June 2024 and the project is initially intended to improve the temporal consistency of monocular depth estimation. Our initial idea is to adopt DUSt3R~\cite{wang2024dust3r} to align estimated depth maps of different frames. Thus, our codes are mainly based on DUSt3R and we incorporate the estimated depth maps in fine-tuning DUSt3R. 

MonST3R~\cite{zhang2024monst3r} is released on arXiv in October 2024, which aims to extend the DUSt3R model on dynamic videos without utilizing monocular depth estimation. Thus, our motivation is different from MonST3R but leads to a similar solution in the end. We find that the flow loss proposed in MonST3R is very important for pose estimation and thus we utilize the flow loss of MonST3R in our implementation. We sincerely thank the authors of DUSt3R and MonST3R for sharing their codes of these two great works.


